\documentclass[sigconf, balance=false]{ACM_Conference_Proceedings_Primary_Article_Template/acmart}
\graphicspath{{ACM_Conference_Proceedings_Primary_Article_Template/}}
\usepackage[most]{tcolorbox}
\usepackage{tabularx}
\usepackage{array}
\usepackage{makecell}
\usepackage{adjustbox}
\usepackage{booktabs}
\usepackage{multirow}
\usepackage{graphicx}
\usepackage{xcolor}
\usepackage{pifont}
\usepackage{ulem}
\usepackage[table]{xcolor}
\usepackage{float}
\usepackage{xurl}

\newcolumntype{L}[1]{%
  >{\raggedright\arraybackslash}p{#1}%
}

\newcolumntype{Y}{%
  >{\raggedright\arraybackslash\sloppy}X%
}

\definecolor{metricgroup}{gray}{0.93}
\newcolumntype{L}[1]{>{\raggedright\arraybackslash}p{#1}}
\newcolumntype{Y}{>{\raggedright\arraybackslash}X}

\newcommand{\cmark}{\ding{51}}
\newcommand{\xmark}{\ding{55}}

\newtcolorbox{cardexample}[1]{
    enhanced,
    breakable,
    width=\columnwidth,
    colback=gray!4,
    colframe=gray!35,
    colbacktitle=gray!55,
    coltitle=white,
    fonttitle=\bfseries\small,
    title={#1},
    boxrule=0.5pt,
    arc=1.5pt,
    left=7pt,
    right=7pt,
    top=5pt,
    bottom=5pt,
    boxsep=0pt,
    before skip=7pt,
    after skip=9pt,
    before upper={\small}
}

\newtcolorbox{exampleprompt}{
    enhanced,
    breakable,
    colback=white,
    colframe=gray!25,
    boxrule=0.4pt,
    arc=1pt,
    left=6pt,
    right=6pt,
    top=5pt,
    bottom=5pt,
    before skip=5pt,
    after skip=5pt,
    fontupper=\small
}

\usepackage{enumitem}
\setlist[itemize]{noitemsep,leftmargin=*,topsep=0pt}
\setlist[enumerate]{noitemsep,leftmargin=*,topsep=0pt}

\AtBeginDocument{%
  }

\setcopyright{none}
\begin{document}

\newcommand{\ync}[1]{{\color{red}{\it [Yaoning: #1]}}}

\newcommand{\yy}[1]{{\color{orange}{#1}}}
\newcommand{\yyc}[1]{{\color{orange}{\it [yeyu: #1]}}}

\title{CARD: Controlled Agentic Reddit Discussions for Credit Card Simulation}
\thanks{Preprint.}

\author{Yaoning Yu}
\affiliation{
  \institution{University of Illinois Urbana-Champaign}
  \country{United States}
}

\author{Kai-Min Chang}
\affiliation{
  \institution{U.S. Bank}
  \country{United States}
}

\author{Ye Yu}
\affiliation{
  \institution{University of Illinois Urbana-Champaign}
  \country{United States}
}

\author{Yi-Chia Wang}
\affiliation{
  \institution{Stanford University}
  \country{United States}
}

\author{Haojing Luo}
\affiliation{
  \institution{Starc.Institute}
  \country{United States}
}

\author{Haohan Wang}
\affiliation{
  \institution{University of Illinois Urbana-Champaign}
  \country{United States}
}
\renewcommand{\shortauthors}{Yu et al.}

\begin{abstract}
Online credit card discussions provide a natural setting for studying how consumers communicate about financial products. Simulating these discussions requires more than just generating individual comments, the generated threads should also match how real users express themselves and interact with others. We introduce CARD, a framework for generating realistic credit card discussion threads. Given a credit card post and its matched real thread, CARD uses non-verbatim guidance on reply structure, comment function, stance, tone, and conversational variation. A planner
organizes these controls, a writer generates the discussion, and a calibration loop updates comments' populations that contribute to differences between the generated and real thread distributions. We evaluate CARD on real Reddit credit card discussions using lexical, semantic, behavioral, and structural metrics. CARD matches the distributions of real credit card discussions better than simulation baselines across multiple LLMs and also demonstrates smaller effect sizes and distribution distances across metrics. These results show that structured planning and targeted revision can generate the realism of simulated credit card discussions.
\end{abstract}


\settopmatter{
  printacmref=false,
  printccs=true,
  printfolios=true
}
\renewcommand\footnotetextcopyrightpermission[1]{}

\maketitle

\fancyhead{}
\pagestyle{plain}

\section{Introduction}




Credit card decisions depend on more than product terms and fees because consumers also rely on information shared by other users when they
compare products, evaluate rewards, and interpret credit card
benefits \citep{agarwal2015regulating, keys2019minimum,
hong2004social, hong2005thy, ru2016credit}. Online communities have become an important source of this information where consumers ask questions, share personal experiences, compare card features, explain product rules, and correct financial claims \citep{thukral2022understanding}. These discussions reveal how consumers interpret, compare, and
reason about credit card products in practice. Unlike product disclosures, they expose practical experiences,
alternative recommendations, and consumer-specific trade-offs that
help consumers understand the same credit card product from
different perspectives.


\noindent
Realistic simulation of these discussions provides a controlled
environment for studying how consumers interpret and communicate
about credit card products from different perspectives. Researchers
can examine how changes to product design, reward structures, fee
policies, or information presentation influence the perspectives
expressed during consumer discussions without requiring large-scale
real-world interventions.


\noindent
Existing financial simulators mainly model borrowing behavior,
trading decisions, and market dynamics
\citep{byrd2019abides, dwarakanath2024abides,
gao2024simulating, li2024econagent}. They do not generate the
consumer discussion threads through which financial products are explained,
questioned, and compared. Moreover, existing social simulators generate
general online interactions or profile-conditioned discussion threads
\citep{yang2411oasis, yukhymenko2024synthetic,
park2023generative}. They do not model how consumers communicate about credit card
products by exchanging recommendations,
sharing personal experience, and
discussing practical trade-offs \citep{yu2026mirobench}.


\noindent
We present \textsc{CARD}, a framework for simulating realistic credit card
discussions. Rather than prescribing the specific content of each
comment, CARD models how consumers communicate about credit card
products. The planner organizes discussion structure and
communication roles, the writer generates discussion content
conditioned on the seed post and local discussion context, and the
metric-guided revision loop calibrates the generated discussion
collection toward the communication patterns observed in real
consumer communities.




\noindent
\textbf{Contributions.}
This paper makes the following contributions:
\begin{itemize}
    \item We formulate consumer credit card discussion simulation and define realism through thread-level communication patterns observed in real consumer communities.

    \item We propose \textsc{CARD}, a framework that models financial discussion structure, communication roles, reasoning patterns, and product-specific issue coverage to generate realistic consumer credit card discussion threads.

    \item We evaluate \textsc{CARD} on real Reddit credit card discussion threads and show that it produces discussion thread distributions that are consistently closer to real consumer discussion threads than existing social simulation baselines.
\end{itemize}
\section{Related Work}

\subsection{Financial Simulation and Consumer Finance}

Agent-based simulation has long been used in finance to study how heterogeneous individual decisions aggregate into market-level outcomes. ABIDES provides a high-fidelity discrete-event environment for financial-market simulation, while ABIDES-Gym extends this environment to reinforcement-learning research \citep{byrd2019abides, amrouni2021abides}. ABIDES-Economist broadens this line of work beyond market microstructure by modeling interactions among households, firms, a government, and a central bank \citep{dwarakanath2024abides}. More recent studies incorporate large language model (LLM) agents to simulate trading decisions, communication, market responses, and behavioral consistency with established financial theories \citep{gao2024simulating, bi2025agent, li2026behavioral, li2024econagent}. Within consumer credit, agent-based models have also been used to study credit card promotion strategies and heterogeneous customer responses \citep{hamill2025agent}.

\noindent A broad consumer-finance literature shows that credit card behavior is shaped by regulation, contract design, rewards, information presentation, and behavioral biases. The Credit Card Accountability Responsibility and Disclosure (CARD) Act examines how restrictions on fees and repricing, together with changes to statement disclosures, affected consumer borrowing costs, repayment information, and access to credit \citep{agarwal2015regulating}. Studies of minimum-payment disclosures show that salient payment anchors can influence repayment amounts and slow debt reduction, while alternative information designs and repayment cues can change these choices \citep{keys2019minimum, hershfield2015dual, hendy2021unsticking, guttman2023weighing}. 

\noindent Other work finds that rewards and introductory product features can affect spending, borrowing, and product selection, and that issuers may design or target contracts around differences in consumer sophistication and behavioral bias \citep{agarwal2010banks, ru2016credit, bursztyn2018status}. Consumers also frequently depart from cost-minimizing repayment and borrowing strategies across multiple cards, including borrowing on higher-cost cards and allocating repayments according to balances rather than interest rates \citep{ponce2017borrowing, stango2016borrowing, gathergood2019individuals}. 

\noindent Together, these studies show that credit card behavior is heterogeneous, context-dependent, and shaped by social and informational cues rather than product terms alone. Existing simulators, however, focus on decisions and market outcomes rather than the discussions through which consumers form these decisions. CARD targets this discussion layer directly: it simulates the online conversations in which credit card advice, experience, and disagreement are exchanged.

\subsection{LLM-Based Social Simulation and Realism Evaluation}

LLM agents have been used to simulate individual behavior, interpersonal interaction, and large-scale online communities. Generative Agents studies believable behavior in a small simulated community \citep{park2023generative}, S$^3$ simulates emotion, attitude, and interaction dynamics over social networks~\cite{gao2023s3}, SOTOPIA evaluates agents in goal-oriented social interactions \citep{zhou2310sotopia}, OASIS models social-media activity involving user networks, content, actions, and recommendation mechanisms \citep{yang2411oasis}, and AgentSociety simulates societal dynamics with over ten thousand LLM-driven agents~\cite{piao2025agentsociety}. Closer to our setting, SynthPAI generates Reddit-style comment threads from profile-conditioned LLM agents~\cite{yukhymenko2024synthetic}. Surveys of this area organize existing systems across individual-, scenario-, and society-level simulation \citep{mou2026individual}.

\noindent A central challenge is whether these simulations reproduce the aggregate properties of real human communities rather than merely generating plausible individual responses. MiroBench evaluates this form of realism by comparing generated and real Reddit discussion threads across linguistic, behavioral, and structural distributions \citep{yu2026mirobench}. Our work builds on this distributional view and develops a generation framework specifically for credit card discussions, where realism requires communication patterns that reflect credit card discussions.
\section{Methodology}

\begin{figure*}[t]
    \centering
    \includegraphics[width=0.75\linewidth]{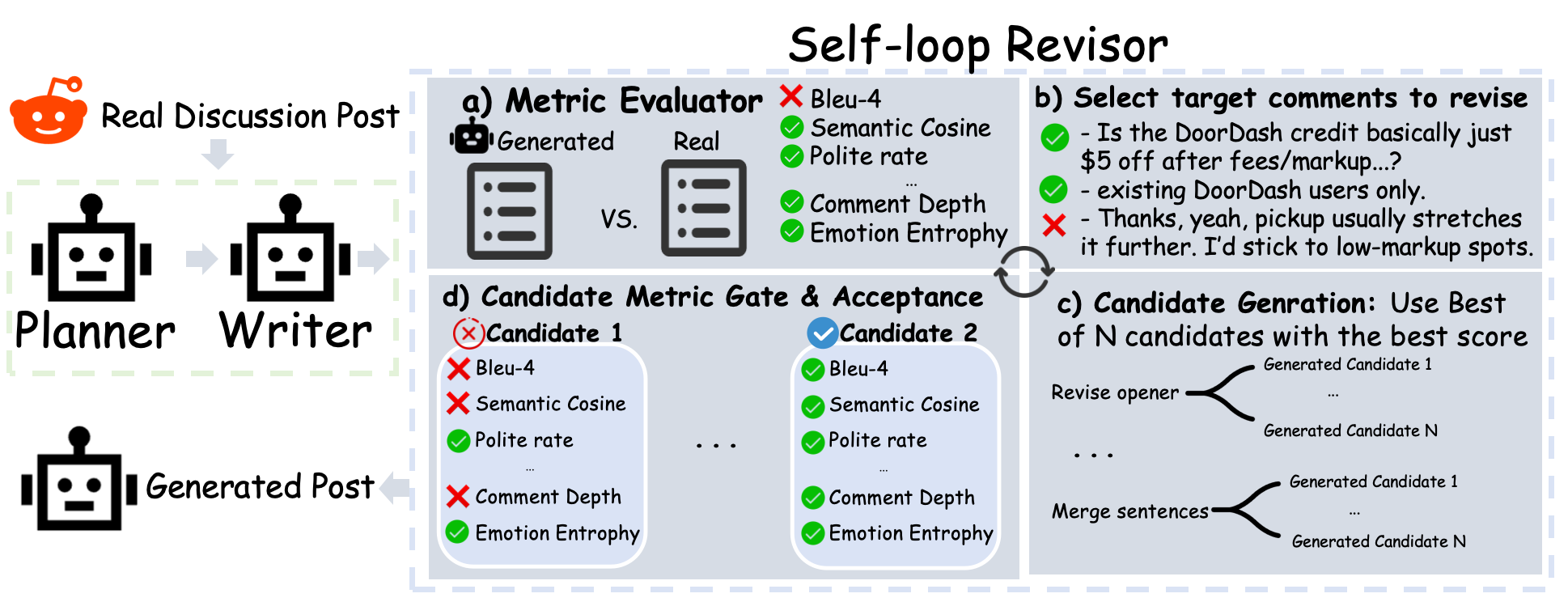}
    \caption{
    Overview of CARD. Given a credit card seed post and its matched
    real discussion.
    The planner controls reply structure,
    financial communication functions, credit card issue types,
    financial claim stances, and reasoning about benefits, costs,
    rules, and trade-offs. The writer follows this plan and a partial
    view of previous comments to generate a new comment. The
    self-looped population revisor then improves the generated discussion
    collection through four steps: (a) the metric evaluator compares
    the generated and real collections across distributional metrics;
    (b) the controller selects comments that contribute to the current
    mismatch; (c) the reviser generates best-of-$N$ candidate rewrites;
    and (d) the metric gate accepts candidates that improve the target
    metric while preserving the planned financial role of each
    comment. The revision loop stops when the target metric is no
    longer significantly different from the real distribution or when
    the maximum number of iterations is reached.
    }
    \label{fig:framework}
\end{figure*}



CARD models how consumers communicate about credit card products, rather than prescribing the specific content of individual comments, through three stages. The discussion planner specifies
how the discussion should develop and assigns a financial role to
each comment. The writer then generates comments that follow these roles
and discuss the planned credit card issues, claims, rules, and
trade-offs. The self-looped calibrator then revises all the generated
discussion thread collection toward the linguistic, behavioral, and
structural patterns observed in real credit card communities.

\subsection{Task Formulation}
CARD represents an online credit card discussion as a discussion
thread rooted at a credit card seed post. We use discussion
to refer to the complete thread, including the seed post, comments,
and reply relations. A post $p$ is the root item that starts a discussion:
$p=(p_{\text{title}},p_{\text{body}}, \allowbreak
p_{\text{prod}},p_{\text{comm}}).$ A comment $c_i$ is a response either to the root post or to another
comment. A discussion thread contains the root post and all
associated comments and reply relations:
\[
T_p =
\left(
p,
\{(c_i,r_i)\}_{i=1}^{n_p}
\right),
\]
where $r_i$ identifies the parent of comment $c_i$, either the root
post or an earlier comment. For a collection of seed posts
$\mathcal{P}$, CARD generates a collection of threads
\(
\hat{\mathcal{T}}
=
\{\hat{T}_p\}_{p\in\mathcal{P}}.
\)

\noindent The seed post defines the credit card product, consumer
question, and financial context. CARD generates the remaining
discussion, including the reply structure, number of comments, and comment text. 
At the discussion level, CARD controls the number of
comments, reply depth, and branch structure. At the comment level,
it controls each comment's financial discussion function, credit
card issue type, financial reasoning move, claim stance, tone,
length, and opening style.

\noindent Across the collection of discussions, CARD aims to reproduce the
population-level communication patterns of real credit card
communities, including lexical and semantic variation, disagreement,
tone, personal stories, emotion, and reply structure.


\subsection{Distributional Matching Objective}

Let $\mathcal{M}=\{\phi_1,\phi_2,\ldots,\phi_K\}$ denote the set of
$K$ evaluation metrics. For a discussion thread $T_p$, let
$\phi_k(T_p)$ denote the value of the $k$-th metric, where
$k\in\{1,\ldots,K\}$. The metrics cover linguistic, behavioral,
and structural aspects, including lexical repetition, semantic
variation, comment length, reply depth, disagreement, tone, story
content, and emotion.
Their full definitions are given in Section~\ref{sec:evaluation}. For a
discussion collection $\mathcal{T}$, define the empirical metric
distribution
\[
\Phi_k(\mathcal{T})
=
\{\phi_k(T):T\in\mathcal{T}\}.
\]
Given a generated discussion thread collection $\hat{\mathcal{T}}$ and
its matched real thread collection $\mathcal{T}^{\mathrm{real}}$, CARD
generates discussions that reduce the discrepancy between
$\Phi_k(\hat{\mathcal{T}})$ and
$\Phi_k(\mathcal{T}^{\mathrm{real}})$
for each metric $\phi_k\in\mathcal{M}$.

\noindent\textbf{Statistical diagnostics.}
We evaluate distributional alignment between generated and real discussions using the Mann--Whitney U and Kolmogorov--Smirnov tests, together with Cliff's $\delta$ and Wasserstein distance. We use these tests as mismatch diagnostics.

\subsection{Framework Overview}

For each credit card seed post, CARD retrieves one matched real
with the same about the same financial scenario. The matched discussion
provides non-verbatim guidance for how consumers organize financial
advice, product explanations, claim corrections, and trade-off
reasoning. It does not provide exact text to copy. The overall process is
\[
A_p = P(p,T_p^{\mathrm{real}}),
\qquad
\tilde{T}_p = W(p,A_p),
\qquad
\hat{T}_p =
R\!\left(
\tilde{T}_p,
E(\tilde{T}_p,T_p^{\mathrm{real}})
\right).
\]
Here, $P$ builds an abstract discussion plan $A_p$ from the seed post $p$
and matched real discussion $T_p^{\mathrm{real}}$. The plan $A_p$ specifies discussion structure
and the financial role of each comment. The writer $W$ generates an
initial discussion $\tilde{T}_p$ that follows this plan based on the seed post $p$. The evaluator
$E$ computes discussion-level metric values, and the reviser $R$
calibrates the generated discussion collection toward the
communication patterns of real credit card communities.
Figure~\ref{fig:framework} presents an overview of the framework.

\subsubsection{\textbf{Discussion planner}}


The planner models recurring communication behaviors in credit card
discussions before any comment is generated. These behaviors include
requesting product advice, recommending or rejecting a card,
explaining reward and eligibility rules, qualifying financial claims,
comparing fees and benefits, and introducing practical trade-offs.
The planner represents these behaviors through discussion-level and
comment-level controls.

\noindent\textbf{Discussion level.}
At the discussion level, CARD plans how a credit card conversation
develops through questions, recommendations, explanations,
corrections, and follow-up replies. It controls the number of
comments, reply depth, branch structure, and the balance between
top-level and nested replies. These quantities are derived from the
matched real discussion and adjusted by global generation
parameters, such as comment count and discussion size.

\begin{table}[t]
\centering
\footnotesize
\setlength{\tabcolsep}{3pt}
\renewcommand{\arraystretch}{1.08}
\caption{Credit card discussion controls used by the CARD
planner.}
\label{tab:planner_controls}
\begin{tabularx}{\columnwidth}
{@{}p{0.07\columnwidth}p{0.34\columnwidth}X@{}}
\toprule
Symbol & Control & Purpose \\
\midrule
$r_i$
& Parent
& Places the comment in the reply tree \\

$d_i$
& Depth
& Controls its structural position \\

$\ell_i$
& Length
& Controls the approximate comment length \\

$f_i$
& Financial discussion function
& Sets the financial role of the comment \\

$s_i$
& Financial claim stance
& Controls support, qualification, or disagreement with a claim \\

$m_i$
& Financial reasoning move
& Specifies how the comment develops a financial issue or claim \\

$q_i$
& Financial issue type
& Identifies the financial aspect of the ongoing discussion to which the comment responds\\

$\tau_i$
& Tone
& Controls how the comment is expressed \\

$o_i$
& Opening style
& Reduces repeated comment openings \\
\bottomrule
\end{tabularx}
\end{table}

\noindent\textbf{Comment level.}
For each planned comment, CARD specifies its communication role
before text generation. The abstract plan is
\[
A_p = \{a_i\}_{i=1}^{\hat{n}},
\]
where $\hat{n}$ is the target number of generated comments. Each
comment plan is represented as
\[
a_i =
\left(
r_i,
d_i,
\ell_i,
f_i,
s_i,
m_i,
q_i,
\tau_i,
o_i
\right).
\]

\noindent Together, these controls define the role of each comment in the
credit card discussion. A comment may request advice, recommend a
product, explain a reward or eligibility rule, qualify a product
claim, correct financial advice, share product experience, or
introduce a cost--benefit trade-off. The financial issue type $q_i$
identifies the credit card issue addressed by the comment, such as
rewards, fees, eligibility, approval, product value, or card usage.
The financial reasoning move $m_i$ specifies how the comment develops
that issue, for example by comparing alternatives, explaining a
rule, adding a practical constraint, or challenging an earlier
claim. These controls specify how each comment participates in the
discussion rather than what it should say. The specific content is
generated by the writer conditioned on the seed post and the local
discussion context. Table~\ref{tab:planner_controls} summarizes this control
set.

\begin{cardexample}{Running Example: Planner Output}

\textbf{Thread structure:}
37 comments; 12 top-level replies; maximum depth 4.

\smallskip
\textbf{Planned comment:}

\begin{tabularx}{\linewidth}{@{}>{\bfseries}lX@{}}
Function:
& Skeptical correction with a practical caveat. \\

Issue:
& Practical value of the DoorDash credit. \\

Reasoning move:
& Introduce fees and markup as a trade-off. \\

Stance:
& Qualify the parent claim. \\

Tone:
& Casual and skeptical. \\

Opening:
& ``Sure, if you ignore the delivery fees \ldots'' \\
\end{tabularx}

\end{cardexample}

\subsubsection{\textbf{Writer}}

The writer generates each comment that executes the planned communication role while remaining grounded in the ongoing credit card discussion initiated by the seed post. For the $i$-th comment, the writer produces
\[
c_i
\sim
P_{\theta}
\left(
c_i
\mid
p,
a_i,
c_{r_i},
H_i
\right),
\]
where $a_i$ is the comment plan, $c_{r_i}$ is the generated parent
comment when applicable, and $H_i$ is the visible portion of the
previously generated comments in the same discussion.

\noindent The writer receives the seed post, its credit card product
context, the comment plan, the generated parent comment, and selected
previous comments. It then generates a new comment that performs the
planned financial role. For example, the comment may recommend a
card, explain a reward rule, compare fees and benefits, qualify a
claim, or introduce a spending constraint. Since comments are
generated in reply order, later comments can extend, qualify,
correct, question, support, or challenge earlier financial claims.

\noindent\textbf{Prompt controls.}
Each field in $a_i$ is prompted into an explicit writer instruction.
The instruction specifies different aspects such as the target length, financial claim stance, tone,
financial issue type, and financial reasoning move.
These controls determine both what financial issue the comment
addresses and how it develops the discussion. The model then
generates new comments conditioned on the seed post and local discussion
context.

\noindent\textbf{Context dropout and variation.}
Consumers rarely read an entire discussion before replying.
CARD therefore varies the visible discussion history during
generation. Context dropout removes part of the previously generated discussion from
the current writer's view, while context variation changes which
earlier comments are emphasized in the prompt. The seed post and the
direct parent comment remain available so that the generated comment
continues to address the current credit card discussion. This design is inspired by dropout in neural network training, where
randomly omitting part of the available representation reduces
co-adaptation and encourages more robust behavior
\citep{srivastava2014dropout}. In CARD, exposing each generated
comment to a partial and slightly different view of the discussion
encourages varied responses and reduces repeated semantic patterns.
It also reflects online discussion behavior, where participants do
not attend to every earlier comment equally.

\begin{cardexample}{Running Example: Writer Prompt}

\smallskip
Write a comment based on the planned instruction and previous
comments.

\smallskip
\textbf{Planned instruction:}
skeptical correction; casual tone; qualify the parent claim; discuss
the pickup-markup trade-off...

...

\smallskip
\textbf{Visible context:}
[a partial view of earlier comments discussing pickup prices]

\smallskip
\textbf{Writing instructions:}
write a natural Reddit-style reply, do not copy previous comments,
and reply directly to the parent.

\end{cardexample}

\subsubsection{\textbf{Self-Looped population calibrator.}}

Individual comments may follow their planned financial roles while
the overall discussion collection still differs from real credit card
communities. CARD therefore applies a self-looped calibration stage to
calibrate collection-level communication patterns after all the discussion threads are generated. It
improves one target metric at a time while treating all other metrics as protected. It also preserves the credit card discussion
plan, such as the seed post, reply tree, financial discussion function, financial claim stance, and financial issue type. This prevents a metric-oriented rewrite from
changing a recommendation into an explanation, 
reversing a financial claim, or dropping a
cost--benefit trade-off. The revision loop continues until neither
statistical test detects a significant difference for the target
metric or the maximum number of revision iterations is reached.

\noindent\textbf{Round diagnosis and protection policy.}
Each revision round starts by comparing the current generated
thread collection with the matched real thread collection using the metric families
in Section~\ref{sec:evaluation}. If statistical tests detect a
significant difference between the generated and real distributions
for a metric, the controller selects that metric as the target metric
$\phi_t$. All other metrics form the protected set
\[
\mathcal{P}_t
=
\mathcal{M}\setminus\{\phi_t\}.
\]
The round is allowed to reduce the mismatch in $\phi_t$, while the
metrics in $\mathcal{P}_t$ must remain within tolerance. The
controller also identifies where the generated--real gap occurs,
such as the upper tail or middle mass, and selects the revision
operations for the round.

\begin{cardexample}{Running Example: Round Diagnosis}

\begin{tabularx}{\linewidth}{@{}Xr@{}}
\textbf{Metric} & \textbf{Result} \\
\hline
Self-BLEU
& \textcolor{red}{\xmark} \\

Semantic cosine similarity
& \textcolor{green!60!black}{\cmark} \\

Disagreement rate
& \textcolor{green!60!black}{\cmark} \\

Politeness rate
& \textcolor{green!60!black}{\cmark} \\

Comment length
& \textcolor{green!60!black}{\cmark} \\

Comment depth
& \textcolor{green!60!black}{\cmark} \\

Emotion entropy
& \textcolor{green!60!black}{\cmark} \\
\end{tabularx}

\smallskip
\textbf{Target metric:} Self-BLEU.

\textbf{Protected metrics:} All remaining metrics.

\end{cardexample}

\noindent\textbf{Thread and comment selection.}
Given a target metric $\phi_t$, CARD first computes a target-metric
gap for each generated thread. For a generated thread $T_p$ and its
matched real thread $T_p^{\mathrm{real}}$, the gap is
\[
g_t(T_p)
=
\left|
\phi_t(T_p)
-
\phi_t(T_p^{\mathrm{real}})
\right|.
\]
Threads with larger $g_t(T_p)$ are selected first because they
contribute more to the current mismatch. Within each selected thread, CARD ranks comments by how strongly they
contribute to the target metric. For example, in Self-BLEU, comment contribution
is estimated from within-thread lexical overlap. CARD compares the
target comment with other comments in the same thread, counts shared
repeated $n$-grams, and detects repeated opener and connector
patterns. Comments with larger contribution to the gap are revised first.

\begin{cardexample}{Running Example: Thread and Comment Selection}

\textcolor{green!60!black}{\cmark}\;
\emph{Is the DoorDash credit basically just \$5 off after fees and
markup?}

\smallskip
\textcolor{green!60!black}{\cmark}\;
\emph{Existing DoorDash users only.}

\smallskip
\textcolor{red}{\xmark}\;
\emph{Thanks, yeah, pickup usually stretches it further. I'd stick
to low-markup spots.}

\smallskip
\textcolor{green!60!black}{\cmark}\;
\emph{Just a clearing delay, not a card glitch.}

\smallskip
The third comment is selected because it contributes most to the
Self-BLEU mismatch.

\end{cardexample}

\noindent\textbf{Candidate generation.}
For each selected comment, CARD first samples candidate rewrites
using a predefined set of metric-specific revision operations. The
controller selects these operations based on the current target
metric and the detected failure pattern. If the predefined operations do not produce an effective revision,
the controller asks an LLM to generate additional revision
instructions. These instructions are conditioned on the current
failure pattern and the revision history, including previously used
instructions and their outcomes. This allows the controller to avoid
repeating ineffective edits and to propose a more targeted revision
strategy for the current comment. If the controller selects $M$ operations and samples N candidates for each operation, CARD evaluates $M \times N$ candidate rewrites for that comment:
\[
\mathcal{C}_i
=
\left\{
c_i^{\prime(m,n)}
\right\}_{m=1,n=1}^{M,N}.
\]
The candidate prompt includes the target comment, its parent comment,
selected earlier comments, and the planned financial discussion
function, claim stance, issue type, and reasoning move. The rewrite
must preserve the credit card content needed by that role, such as a
product recommendation, reward rule, eligibility condition, fee
comparison, or spending trade-off.

\begin{cardexample}{Running Example: Candidate Generation}

\textbf{Target comment:}

\emph{Thanks, yeah, pickup usually stretches it further. I'd stick
to low-markup spots.}

\smallskip
\textbf{1. Revise opener}

\begin{enumerate}
    \setlength{\itemsep}{1pt}
    \setlength{\topsep}{2pt}
    \item \emph{Yeah, pickup usually costs more. I'd stick to
    low-markup spots.}
    
    ...
\end{enumerate}

\textbf{2. Change wording}

\begin{enumerate}
    \setlength{\itemsep}{1pt}
    \setlength{\topsep}{2pt}
    \item \emph{Pickup usually adds even more. I'd go with
    low-markup places.}
    
    ...
\end{enumerate}

\textbf{3. Merge sentences}

\begin{enumerate}
    \setlength{\itemsep}{1pt}
    \setlength{\topsep}{2pt}
    \item \emph{Since pickup usually costs more, I'd stick to
    low-markup spots.}
    
    ...
\end{enumerate}

\end{cardexample}


\noindent\textbf{Candidate scoring and acceptance.}
For each candidate rewrite $c_i'\in\mathcal{C}_i$, CARD forms a
candidate thread $T_i'$ by replacing only $c_i$ with $c_i'$. The
candidate receives a target-gap reduction score:
\[
R_t(c_i')
=
g_t(T^{(m)})
-
g_t(T_i'),
\]
where $T^{(m)}$ is the current thread before the edit. A positive
$R_t(c_i')$ indicates that the candidate reduces the mismatch on the
target metric. A candidate is locally accepted when
\[
R_t(c_i')>\epsilon_t,
\qquad
\operatorname{Preserve}(a_i,c_i')=1,
\]
where $\epsilon_t$ is the minimum required reduction in the target
metric gap. The preservation check requires the rewrite to retain the
planned comment attributes in $a_i$, including the financial discussion
function, claim stance, issue type, reasoning move, approximate length,
and reply target. This check preserves the planned financial role and
credit card content of the comment.

\noindent CARD may also apply metric-specific local gates. For example, the
Self-BLEU reviser checks whether the rewritten comment reduces its
lexical overlap with other comments in the same discussion. Among
locally accepted candidates, CARD selects the rewrite with the largest
target-gap reduction.
If no candidate satisfies the acceptance conditions, the original
comment is left unchanged.
\begin{cardexample}{Running Example: Candidate Acceptance}

\begin{minipage}[t]{0.47\linewidth}
\textbf{Candidate 1}

\smallskip
\emph{Yeah, pickup is still expensive. Use cheaper places.}

\smallskip
\textcolor{red}{\xmark}\; Self-BLEU

\textcolor{red}{\xmark}\; Semantic cosine

\textcolor{green!60!black}{\cmark}\; Disagreement

\textcolor{green!60!black}{\cmark}\; Politeness

\textcolor{green!60!black}{\cmark}\; Length

\textcolor{red}{\xmark}\; Emotion

\smallskip
\textbf{Rejected}
\end{minipage}
\hfill
\begin{minipage}[t]{0.47\linewidth}
\textbf{Candidate 2}

\smallskip
\emph{Since pickup often raises the price, I'd stick to restaurants
with low markups.}

\smallskip
\textcolor{green!60!black}{\cmark}\; Self-BLEU

\textcolor{green!60!black}{\cmark}\; Semantic cosine

\textcolor{green!60!black}{\cmark}\; Disagreement

\textcolor{green!60!black}{\cmark}\; Politeness

\textcolor{green!60!black}{\cmark}\; Length

\textcolor{green!60!black}{\cmark}\; Emotion

\smallskip
\textbf{Accepted}
\end{minipage}

\end{cardexample}

\noindent\textbf{Applying candidate edits.}
Each accepted rewrite replaces the original comment at the same
position in its thread. The root post, parent relation, and all
unrevised comments remain unchanged. After all selected comments
have been processed, the updated threads form a proposed thread
collection for the current round:
\(
\hat{\mathcal{T}}_{\mathrm{prop}}^{(r+1)}.
\)
CARD then recomputes every evaluation metric over the complete
proposed collection. Local candidate scores therefore determine
which edits are proposed, while collection-level statistics
determine whether the entire revision round is retained.

\noindent\textbf{Round-level acceptance and rollback.}
CARD maintains the most recent accepted thread collection
$\hat{\mathcal{T}}^{(r)}$. In round $r+1$, the reviser produces a
proposed collection
$\hat{\mathcal{T}}_{\mathrm{prop}}^{(r+1)}$. CARD then rescores the
complete proposed collection and compares its metric quality with
that of the most recent accepted collection. The proposed round is accepted only if it improves the target metric
and does not cause any protected metric to degrade beyond tolerance. If accepted, the proposed collection becomes
the input to the next round; otherwise, CARD rolls back to the most
recent accepted collection:
\[
\hat{\mathcal{T}}^{(r+1)}
=
\begin{cases}
\hat{\mathcal{T}}_{\mathrm{prop}}^{(r+1)},
& \text{if the round is accepted},\\
\hat{\mathcal{T}}^{(r)},
& \text{otherwise}.
\end{cases}
\]
For each target metric, the revision loop continues until either no significant difference is detected or the maximum of seven revision iterations is reached. The controller then selects the next unfinished mismatch metric and start from round diagnosis and protection policy through the self-revision loop. The overall revision process terminates only after every mismatch metric has either passed the statistical test or exhausted its revision budget. CARD finally returns the most recently accepted collection.

\begin{table*}[!t]
\centering
\footnotesize
\setlength{\tabcolsep}{1.45pt}
\renewcommand{\arraystretch}{1.10}

\begin{adjustbox}{max width=\textwidth}
\begin{tabular}{@{}ll*{12}{cc}c@{}}
\toprule

\multirow{2}{*}{Method}
& \multicolumn{1}{c}{\multirow{2}{*}{Model}}
& \multicolumn{2}{c}{Self-BLEU}
& \multicolumn{2}{c}{Self-BERT}
& \multicolumn{2}{c}{Sem. Cos.}
& \multicolumn{2}{c}{Hard Dis.}
& \multicolumn{2}{c}{Polite}
& \multicolumn{2}{c}{Impolite}
& \multicolumn{2}{c}{Neutral}
& \multicolumn{2}{c}{Len. CV}
& \multicolumn{2}{c}{Depth}
& \multicolumn{2}{c}{Virality}
& \multicolumn{2}{c}{Story}
& \multicolumn{2}{c}{Emotion}
& \multirow{2}{*}{\makecell{Non-rej.\\(\%)}} \\

\cmidrule(lr){3-4}
\cmidrule(lr){5-6}
\cmidrule(lr){7-8}
\cmidrule(lr){9-10}
\cmidrule(lr){11-12}
\cmidrule(lr){13-14}
\cmidrule(lr){15-16}
\cmidrule(lr){17-18}
\cmidrule(lr){19-20}
\cmidrule(lr){21-22}
\cmidrule(lr){23-24}
\cmidrule(lr){25-26}

& & MWU & KS
& MWU & KS
& MWU & KS
& MWU & KS
& MWU & KS
& MWU & KS
& MWU & KS
& MWU & KS
& MWU & KS
& MWU & KS
& MWU & KS
& MWU & KS
& \\

\midrule

\multirow{4}{*}{OASIS}
& \texttt{gemini-2.5-flash}
& $<\!0.001$ & $<\!0.001$
& $<\!0.001$ & $<\!0.001$
& $<\!0.001$ & $<\!0.001$
& $<\!0.001$ & $<\!0.001$
& $<\!0.001$ & $<\!0.001$
& $<\!0.001$ & $<\!0.001$
& $<\!0.001$ & $<\!0.001$
& $<\!0.001$ & $<\!0.001$
& $<\!0.001$ & $<\!0.001$
& $<\!0.001$ & $<\!0.001$
& $<\!0.001$ & $<\!0.001$
& $<\!0.001$ & $<\!0.001$
& 0.0 \\

& \texttt{deepseek-v4-flash}
& $<\!0.001$ & $<\!0.001$
& $<\!0.001$ & $<\!0.001$
& $<\!0.001$ & $<\!0.001$
& $<\!0.001$ & $<\!0.001$
& $<\!0.001$ & $<\!0.001$
& $<\!0.001$ & $<\!0.001$
& $<\!0.001$ & $<\!0.001$
& $<\!0.001$ & $<\!0.001$
& $<\!0.001$ & $<\!0.001$
& $<\!0.001$ & $<\!0.001$
& $<\!0.001$ & $<\!0.001$
& 0.037 & \textbf{0.073}
& 4.2 \\

& \texttt{gpt-4o-mini}
& $<\!0.001$ & $<\!0.001$
& $<\!0.001$ & $<\!0.001$
& $<\!0.001$ & $<\!0.001$
& 0.001 & $<\!0.001$
& $<\!0.001$ & $<\!0.001$
& $<\!0.001$ & $<\!0.001$
& $<\!0.001$ & $<\!0.001$
& $<\!0.001$ & $<\!0.001$
& $<\!0.001$ & $<\!0.001$
& $<\!0.001$ & $<\!0.001$
& $<\!0.001$ & $<\!0.001$
& 0.048 & \textbf{0.053}
& 4.2 \\

& \texttt{gpt-5.4-mini}
& $<\!0.001$ & $<\!0.001$
& $<\!0.001$ & $<\!0.001$
& $<\!0.001$ & $<\!0.001$
& 0.039 & 0.020
& $<\!0.001$ & $<\!0.001$
& $<\!0.001$ & $<\!0.001$
& $<\!0.001$ & $<\!0.001$
& $<\!0.001$ & $<\!0.001$
& $<\!0.001$ & $<\!0.001$
& $<\!0.001$ & $<\!0.001$
& $<\!0.001$ & $<\!0.001$
& $<\!0.001$ & $<\!0.001$
& 0.0 \\
\midrule

\multirow{4}{*}{SynthPAI}
& \texttt{gemini-2.5-flash}
& $<\!0.001$ & $<\!0.001$
& $<\!0.001$ & $<\!0.001$
& \textbf{0.186} & \textbf{0.073}
& $<\!0.001$ & $<\!0.001$
& $<\!0.001$ & $<\!0.001$
& $<\!0.001$ & $<\!0.001$
& \textbf{0.054} & \textbf{0.098}
& $<\!0.001$ & $<\!0.001$
& \textbf{0.102} & 0.004
& \textbf{0.197} & 0.013
& $<\!0.001$ & $<\!0.001$
& $<\!0.001$ & $<\!0.001$
& 25.0 \\

& \texttt{deepseek-v4-flash}
& $<\!0.001$ & $<\!0.001$
& $<\!0.001$ & $<\!0.001$
& 0.034 & \textbf{0.169}
& $<\!0.001$ & $<\!0.001$
& $<\!0.001$ & $<\!0.001$
& $<\!0.001$ & $<\!0.001$
& $<\!0.001$ & $<\!0.001$
& \textbf{0.495} & \textbf{0.129}
& \textbf{0.868} & \textbf{0.073}
& \textbf{0.581} & \textbf{0.073}
& \textbf{0.260} & \textbf{0.073}
& $<\!0.001$ & $<\!0.001$
& 37.5 \\

& \texttt{gpt-4o-mini}
& $<\!0.001$ & $<\!0.001$
& $<\!0.001$ & $<\!0.001$
& $<\!0.001$ & $<\!0.001$
& $<\!0.001$ & 0.006
& $<\!0.001$ & $<\!0.001$
& $<\!0.001$ & $<\!0.001$
& $<\!0.001$ & $<\!0.001$
& $<\!0.001$ & $<\!0.001$
& \textbf{0.896} & \textbf{0.073}
& \textbf{0.881} & 0.039
& $<\!0.001$ & $<\!0.001$
& \textbf{0.731} & \textbf{0.129}
& 20.8 \\

& \texttt{gpt-5.4-mini}
& $<\!0.001$ & $<\!0.001$
& $<\!0.001$ & $<\!0.001$
& \textbf{0.858} & \textbf{0.276}
& $<\!0.001$ & $<\!0.001$
& $<\!0.001$ & $<\!0.001$
& $<\!0.001$ & $<\!0.001$
& 0.002 & $<\!0.001$
& $<\!0.001$ & $<\!0.001$
& $<\!0.001$ & $<\!0.001$
& $<\!0.001$ & $<\!0.001$
& $<\!0.001$ & $<\!0.001$
& $<\!0.001$ & $<\!0.001$
& 8.3 \\

\midrule

\multirow{4}{*}{\textbf{CARD}}
& \texttt{gemini-2.5-flash}
& \textbf{0.159} & \textbf{0.059}
& $<\!0.001$ & $<\!0.001$
& \textbf{0.917} & \textbf{0.443}
& \textbf{0.233} & \textbf{0.058}
& \textbf{0.508} & \textbf{0.725}
& $<\!0.001$ & $<\!0.001$
& \textbf{0.361} & \textbf{0.532}
& \textbf{0.392} & \textbf{0.817}
& \textbf{0.094} & \textbf{0.291}
& \textbf{0.053} & \textbf{0.291}
& \textbf{0.979} & \textbf{0.984}
& \textbf{0.063} & \textbf{0.180}
& \textbf{83.3} \\

& \texttt{deepseek-v4-flash}
& \textbf{0.381} & \textbf{0.106}
& \textbf{0.973} & \textbf{0.725}
& \textbf{0.266} & \textbf{0.443}
& \textbf{0.458} & \textbf{0.725}
& \textbf{0.245} & \textbf{0.817}
& \textbf{0.489} & \textbf{0.817}
& \textbf{0.805} & \textbf{0.951}
& \textbf{0.537} & \textbf{0.628}
& \textbf{0.580} & \textbf{0.951}
& \textbf{0.452} & \textbf{0.894}
& 0.019 & 0.011
& \textbf{0.792} & \textbf{0.725}
& \textbf{91.7} \\

& \texttt{gpt-4o-mini}
& 0.020 & 0.011
& $<\!0.001$ & $<\!0.001$
& 0.014 & 0.015
& 0.024 & 0.001
& \textbf{0.344} & \textbf{0.231}
& \textbf{0.081} & \textbf{0.231}
& 0.003 & 0.022
& 0.028 & 0.005
& \textbf{0.079} & \textbf{0.180}
& \textbf{0.057} & \textbf{0.139}
& \textbf{0.149} & \textbf{0.291}
& $<\!0.001$ & $<\!0.001$
& \textbf{41.7} \\

& \texttt{gpt-5.4-mini}
& \textbf{0.307} & \textbf{0.139}
& \textbf{0.323} & \textbf{0.362}
& \textbf{0.373} & \textbf{0.532}
& \textbf{0.848} & \textbf{0.997}
& 0.006 & 0.031
& \textbf{0.770} & \textbf{0.997}
& \textbf{0.185} & \textbf{0.725}
& \textbf{0.393} & \textbf{0.291}
& \textbf{0.258} & \textbf{0.816}
& \textbf{0.134} & \textbf{0.532}
& \textbf{0.260} & \textbf{0.291}
& \textbf{0.892} & \textbf{0.816}
& \textbf{91.7} \\
\bottomrule
\end{tabular}
\end{adjustbox}
\caption{Metric-level statistical comparison across methods and models. Each metric reports Mann--Whitney U and Kolmogorov--Smirnov $p$-values. Values with $p>0.05$ are bolded, indicating that the corresponding test does not detect a statistically significant difference between the generated and real distributions (higher is better). Non-rej. (\%) reports the percentage of metric--test pairs for which the null hypothesis is not rejected at $\alpha=0.05$ across all 12 metrics.
}
\label{tab:pvalue_results}
\end{table*}

\section{Experiment}
\subsection{Baselines}

We compare CARD with OASIS and SynthPAI, two LLM-based frameworks
that support social simulation, across four LLMs. \textbf{OASIS}
\citep{yang2411oasis} is a general and scalable social-media
simulator that models user profiles, social networks, user actions,
and recommendation mechanisms. It is designed to study large-scale
social phenomena, including information diffusion, group
polarization, and herd behavior. \textbf{SynthPAI}
\citep{yukhymenko2024synthetic} uses profile-conditioned LLM agents
to generate Reddit-style comments for personal attribute inference.

\begin{table}[t!]
\centering
\scriptsize
\setlength{\tabcolsep}{6pt}
\renewcommand{\arraystretch}{1.08}
\begin{tabular}{llr}
\toprule
Method & Model & Comments \\
\midrule

\multirow{4}{*}{OASIS}
& \texttt{gemini-2.5-flash}   & 2,438 \\
& \texttt{deepseek-v4-flash}  & 935   \\
& \texttt{gpt-4o-mini}        & 1,700 \\
& \texttt{gpt-5.4-mini}       & 1,176 \\
\midrule

\multirow{4}{*}{SynthPAI}
& \texttt{gemini-2.5-flash}   & 2,175 \\
& \texttt{deepseek-v4-flash}  & 1,508 \\
& \texttt{gpt-4o-mini}        & 1,694 \\
& \texttt{gpt-5.4-mini}       & 5,060 \\
\midrule

\multirow{4}{*}{\textbf{CARD}}
& \texttt{gemini-2.5-flash}   & 2,351    \\
& \texttt{deepseek-v4-flash}  & 2,466    \\
& \texttt{gpt-4o-mini}        & 2,574 \\
& \texttt{gpt-5.4-mini}       & 2,674 \\
\bottomrule
\end{tabular}
\caption{
Number of generated comments across methods and models.
}
\label{tab:generation_statistics}
\end{table}

\subsection{Evaluation Metrics}
\label{sec:evaluation}

\begin{table*}[t]
\centering
\scriptsize
\setlength{\tabcolsep}{2.8pt}
\renewcommand{\arraystretch}{1.08}

\resizebox{\textwidth}{!}{%
\begin{tabular}{lccccc|ccc|cc}
\toprule

\multirow{2}{*}{\textbf{Model}}
&
\multicolumn{5}{c|}{\textbf{Revision Process}}
&
\multicolumn{3}{c|}{\textbf{Revision Outcome}}
&
\multicolumn{2}{c}{\textbf{Candidate Rewrites}}
\\

\cmidrule(lr){2-6}
\cmidrule(lr){7-9}
\cmidrule(lr){10-11}

&
\makecell{\textbf{Total Accepted}\\\textbf{Rounds}}

&
\makecell{\textbf{Diversity}}

&
\makecell{\textbf{Tone}}

&
\makecell{\textbf{Story}}

&
\makecell{\textbf{Structure}}

&
\makecell{\textbf{Edited}\\\textbf{Threads}}

&
\makecell{\textbf{Accepted}\\\textbf{Comments}}

&
\makecell{\textbf{Overall Passes}\\\textbf{Before $\rightarrow$ After}}

&
\makecell{\textbf{Generated}\\\textbf{Rewrites}}

&
\makecell{\textbf{Avg. Rewrites}\\\textbf{/ Comment}}

\\

\midrule

\texttt{gpt-4o-mini}
&
7 / 21 (33.3\%)
&
7 / 7 (100\%)
&
0 / 7 (0.0\%)
&
(No revision)
&
0 / 7 (0.0\%)
&
663 / 1267 (52.3\%)
&
1022 / 7545 (13.5\%)
&
5 / 12 $\rightarrow$ 5 / 12 (41.7\% $\rightarrow$ 41.7\%) 
&
121,299
&
16.1
\\

\texttt{gpt-5.4-mini}
&
10 / 13 (76.92\%)
&
6 / 6 (100\%)
&
4 / 7 (57.1\%)
&
(No revision)
&
(No revision)
&
645 / 971 (66.4\%)
&
1536 / 4331 (35.5\%)
&
7 / 12 $\rightarrow$ 11 / 12 (58.3\% $\rightarrow$ 91.7\%) 
&
89,639
&
20.7
\\

\texttt{DeepSeek V4 Flash}
&
11 / 11 (100\%)
&
1 / 1 (100\%)
&
2 / 2 (100\%)
&
7 / 7 (100\%)
&
1 / 1 (100\%)
&
300 / 580 (51.7\%)
&
456 / 1742 (26.2\%)
&
6 / 12 $\rightarrow$ 11 / 12 (50.0\% $\rightarrow$ 91.7\%) 
&
26,770
&
15.4
\\

\texttt{Gemini 2.5 Flash}
&
7 / 14 (50.0\%)
&
6 / 7 (85.7\%)
&
1 / 7 (14.3\%)
&
(No revision)
&
(No revision)
&
402 / 737 (54.5\%)
&
579 / 3697 (15.7\%)
&
6 / 12 $\rightarrow$ 10 / 12 (50.0\% $\rightarrow$ 83.3\%)
&
47,046
&
12.7
\\

\bottomrule
\end{tabular}%
}
\caption{
Summary of CARD's self-loop revision process. Revision Process reports accepted over attempted revision rounds for each revision stage. Overall Passes compares the number of target metrics satisfying the statistical criteria before and after self-loop revision.
}

\label{tab:selfloop_process_summary}
\end{table*}

We follow the MiroBench \citep{yu2026mirobench} credit-card thread-level evaluation protocol.
All methods are evaluated on the same set of 150 seed posts,
with each method generating one discussion thread for each seed post.
Each generated thread is paired with its matched real Reddit discussion,
and we compute the same set of metrics for all generated-real pairs.We then compare each metric distribution across
the generated and real thread collections.
Table~\ref{tab:evaluation_metrics} summarizes what each metric
measures.
Higher Self-BLEU, Self-BERTScore, and mean semantic cosine values
generally indicate greater lexical or semantic similarity among
comments. For the remaining metrics, there is no fixed
direction; the goal is to match the corresponding real distribution.

\begin{table}[H]
\centering
\footnotesize
\setlength{\tabcolsep}{4.5pt}
\renewcommand{\arraystretch}{1.12}

\caption{Thread-level metrics used to compare generated and real
credit card discussion threads.}
\label{tab:evaluation_metrics}

\begin{tabularx}{\columnwidth}
{@{}L{0.34\columnwidth}Y@{}}
\toprule
\textbf{Metric} & \textbf{What it measures} \\
\midrule

\rowcolor{metricgroup}
\multicolumn{2}{@{}l}{\hspace{4pt}\textbf{Lexical and semantic diversity}} \\

Self-BLEU
& Lexical repetition among comments in a thread. \\

Self-BERTScore
& Contextual semantic similarity among comments. \\

Mean semantic cosine
& Mean embedding similarity among comments. \\

\addlinespace[2pt]
\rowcolor{metricgroup}
\multicolumn{2}{@{}l}{\hspace{4pt}\textbf{Tone and disagreement}} \\

Hard disagreement rate
& Frequency of direct disagreement between comments. \\

Politeness rate
& Proportion of comments classified as polite. \\

Impoliteness rate
& Proportion of comments classified as impolite. \\

Neutral rate
& Proportion of comments classified as neutral. \\

\addlinespace[2pt]
\rowcolor{metricgroup}
\multicolumn{2}{@{}l}{\hspace{4pt}\textbf{Structure and length}} \\

Length CV
& Variation in comment length within a thread. \\

Average reply depth
& Average depth of comments in the reply tree. \\

Structural virality
& How widely interactions spread across the reply tree. \\

\addlinespace[2pt]
\rowcolor{metricgroup}
\multicolumn{2}{@{}l}{\hspace{4pt}\textbf{Story and emotion}} \\

Mean story probability
& Presence of personal experience or narrative detail. \\

Emotion entropy
& Variation in emotional tone within a thread. \\

\bottomrule
\end{tabularx}
\end{table}

\noindent\textbf{Statistical comparison.}
For each metric, we compare the generated and real thread-level
distributions using the Mann--Whitney U test and the
Kolmogorov--Smirnov test. The Mann--Whitney U test measures whether
one distribution tends to have larger values than the other, while
the Kolmogorov--Smirnov test measures differences in the overall
distribution shape. A result with $p-value>0.05$ means that the test does
not detect a statistically significant difference at
$\alpha=0.05$; it does not establish that the two distributions are
equivalent. We also report the absolute Cliff’s $\delta$ and Wasserstein distance as measures of the magnitude of the distributional difference, where smaller values indicate better matching.

\begin{figure}[t!]
    \centering
    \includegraphics[width=1.0\columnwidth]{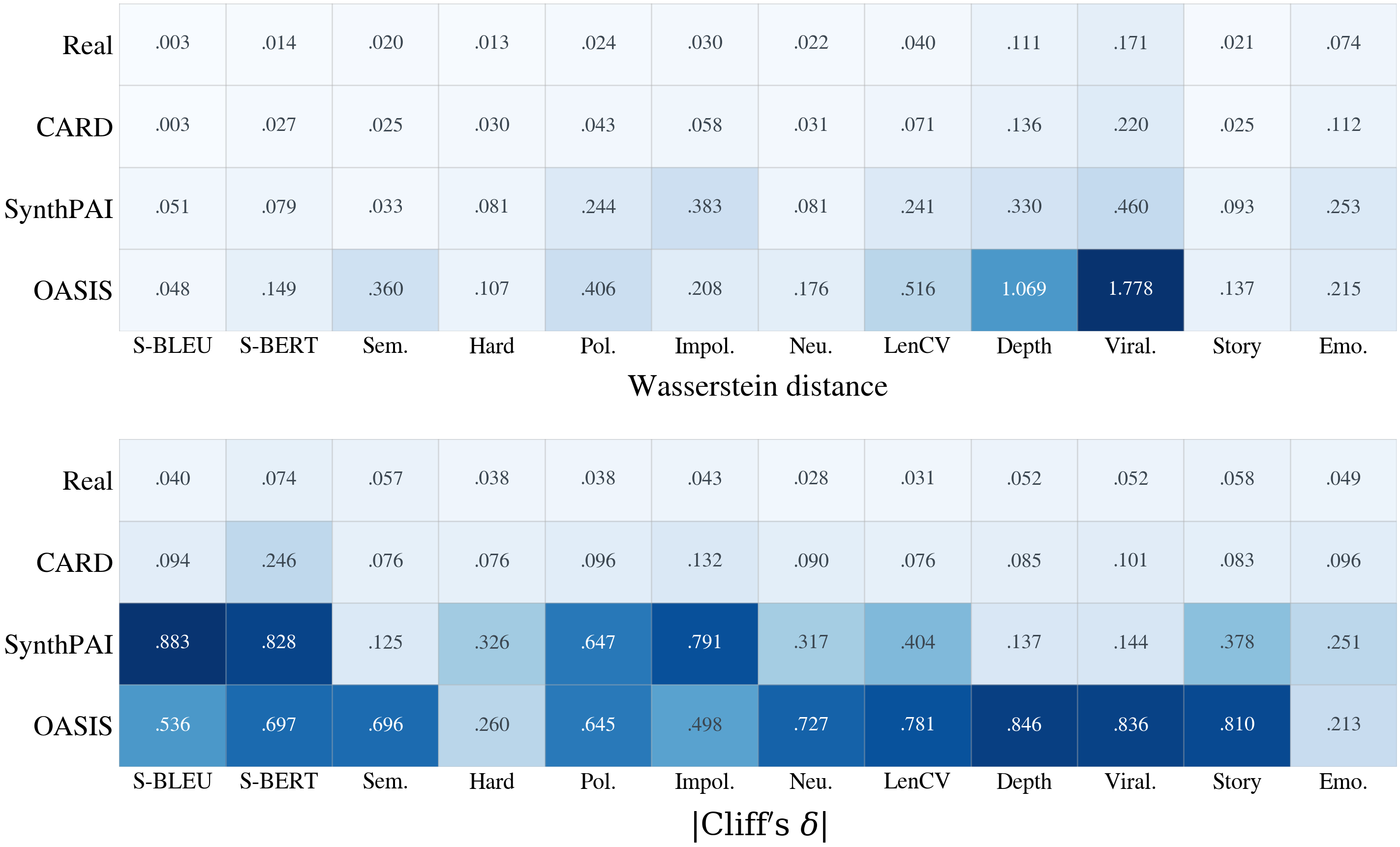}
    \caption{Distributional agreement with real credit-card discussion threads.Lower values indicate closer agreement. Real denotes the real-vs.-real reference averaged over 10 repeated 150-vs.-150 samplings; each method row reports the average distance across four backbone models.}
    \label{fig:distribution_heatmap}
\end{figure}

\subsection{Results Analysis}

\noindent\textbf{Statistical results.}
Since CARD uses statistical tests to guide self-loop revision, we first
evaluate whether the generated distributions become statistically
indistinguishable from the real distributions. Table~\ref{tab:pvalue_results}
reports the Mann--Whitney U and Kolmogorov--Smirnov test results.
CARD with \texttt{gpt-5.4-mini} and \texttt{deepseek-v4-flash} achieves the highest non-rejection
rate, with 91.7\% of the metric--test pairs having $p>0.05$. In
comparison, the strongest SynthPAI configuration reaches 37.5\%,
while OASIS remains at or below 4.2\%. CARD with
\texttt{gpt-5.4-mini} passes both statistical tests on most lexical,
semantic, behavioral, and structural metrics, leaving politeness as
Passing a statistical test, however, does not indicate how close two
distributions actually are. We therefore also report effect sizes and
distribution distances to measure the magnitude of the remaining
difference.

\noindent\textbf{Distributional distance.}
Figure~\ref{fig:distribution_heatmap} complements the statistical
tests by showing the remaining distributional differences measured by
absolute Cliff's $\delta$ and Wasserstein distance. Across almost all
metrics, CARD consistently produces smaller effect sizes and shorter
distributional distances than OASIS and SynthPAI. The improvements
are particularly clear for lexical diversity, semantic variation,
discussion structure, disagreement, story content, and emotion. These
results show that CARD not only satisfies the statistical criteria
more frequently, but also generates discussion distributions that are
quantitatively closer to real credit card discussions.

\begin{figure}[t]
\centering
\includegraphics[width=\linewidth]{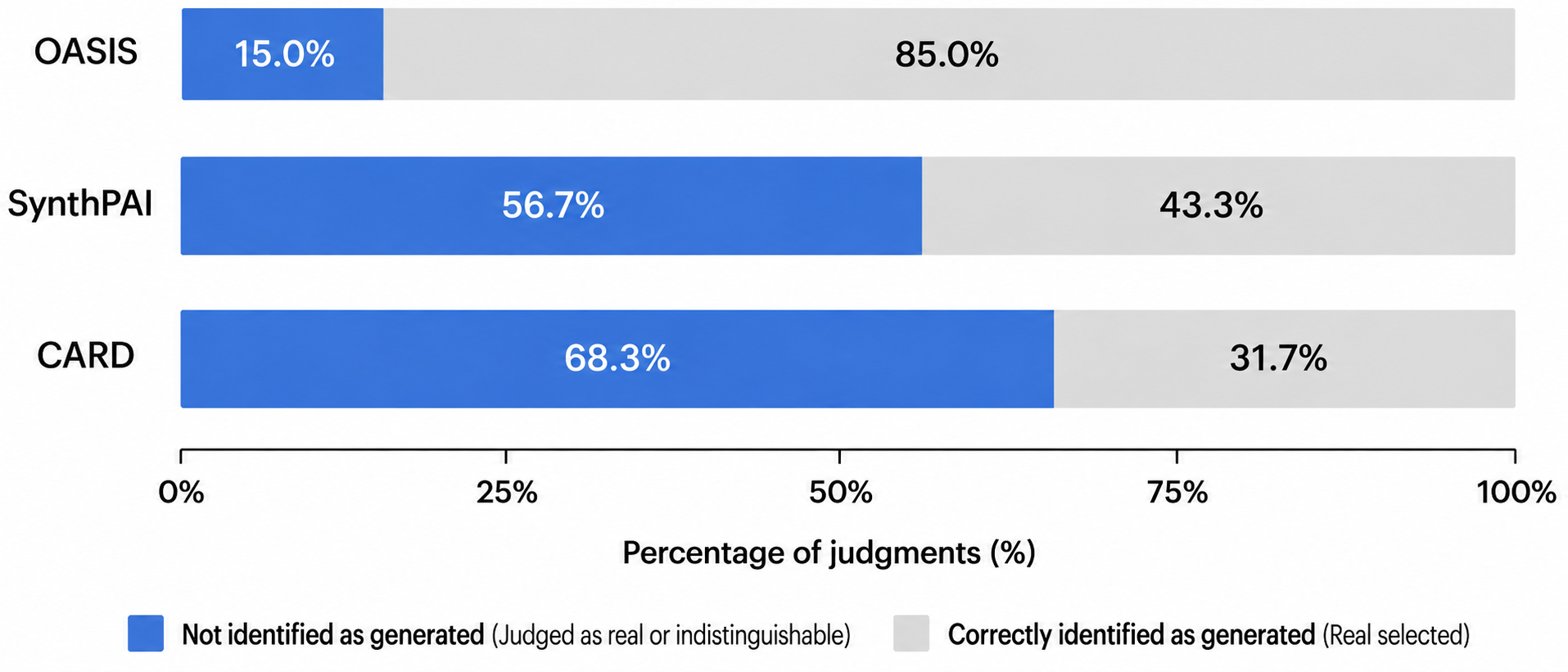}
\caption{
Results of the blind Turing test over 20 discussion pairs for each method. The blue portion is the percentage of judgments in which the generated discussion was not identified as human-written or indistinguishable from the real discussion (higher is better). The gray portion shows the percentage of judgments in which the generated discussion was correctly identified as LLM-generated.
}
\label{fig:turing_test}
\end{figure}

\noindent\textbf{Human evaluation.}
We further evaluated discussion realism using a blind turing test.
For each comparison, annotators were shown one real Reddit discussion
and one generated discussion in random order and asked to identify
which discussion was written by humans, with a third option indicating
that the two discussions could not be distinguished.
We compared CARD against OASIS and SynthPAI using 20 matched discussion
pairs for each method. Five annotators participated in the study. Each annotator evaluated 36 discussion pairs, for a total of 72 unique discussion pairs across all methods. Figure~\ref{fig:turing_test} reports the percentage of judgments in
which the generated discussion was not identified as LLM-generated
(i.e., judged as real or marked as indistinguishable) and the
percentage of correct identifications. CARD achieves the highest
proportion of generated discussions that were not identified,
substantially outperforming both OASIS and SynthPAI.

\noindent\textbf{Effectiveness of self-loop calibration.}
The statistical improvements above are primarily achieved through
CARD's self-loop revision process.
Table~\ref{tab:selfloop_process_summary} shows that self-loop
revision substantially improves distributional alignment across
revision models. Three of the four LLMs increase the number of
metrics that pass the statistical tests, although the degree of
improvement varies considerably across models. \texttt{DeepSeek-V4-Flash} provides the most efficient
revision behavior. It retains all attempted revision rounds and
improves the overall pass count from 6 to 11 metrics while generating
only 26,770 candidate rewrites. Compared with gpt-based revisers, it
requires fewer candidate generations and achieves a higher acceptance
rate of selected comments (26.2\%), suggesting that it more often
produces rewrites that satisfy both target improvement and protection
constraints.



\noindent \texttt{GPT-5.4-mini} also improves the overall pass count
from 7 to 11 metrics and achieves the highest edited-thread rate
(66.4\%) and accepted-comment rate (35.5\%). However, it requires a
larger candidate search, generating 89,639 rewrites with 20.7
rewrites per selected comment. This indicates that GPT-5.4-mini often needs more candidate exploration
before satisfying the acceptance criteria. \texttt{GPT-4o-mini} generates the largest number of
candidate rewrites (121,299), but improves none of the overall passing
metrics. Its accepted-comment rate is only 13.5\%, and all tone and
structure revision rounds fail to be retained. 

\section{Conclusion}

We introduce CARD, a framework for simulating credit card discussion
threads that match how consumers communicate and interact in real
online communities. CARD combines discussion thread planning,
sequential comment generation, and metric-guided revision to control
thread structure, discussion functions, tone, disagreement, and other
communication patterns without copying the content of matched real
threads. Experiments on real Reddit credit card discussion threads show that
CARD reduces linguistic, behavioral, and structural distributional
gaps compared with general social-simulation baselines. These results
suggest that explicit planning and collection-level revision provide
a useful basis for realistic consumer-finance discussion thread simulation.
\section{Future Work}

Future work can extend CARD in two main directions. First, the
framework can be adapted to other financial and non-financial
discussion domains by defining domain-specific discussion functions
and evaluation metrics. Second, the current framework focuses on how
people communicate rather than whether generated threads match the
specific content of real discussions. Future versions could add
content-level objectives, including topic coverage, claim alignment,
and factual consistency, while preserving the linguistic,
behavioral, and structural realism studied in this work.

\bibliographystyle{ACM_Conference_Proceedings_Primary_Article_Template/ACM-Reference-Format}
\bibliography{sections/reference}




\end{document}